\documentclass[a4paper, oneside, 11pt]{article}

\usepackage{graphicx}
\usepackage{amsmath}
\usepackage{amsfonts}

\usepackage{longtable}
\usepackage{booktabs}

\newcommand{\cf}{cf.\,}
\newcommand{\ie}{i.e.\,}

\newcommand{\st}[1]{\mathbb{#1}}
\newcommand{\tn}[1]{\text{#1}}

\begin{document}

\title{Cost-Sensitive Learning \\ for Predictive Maintenance}

\author{
Stephan Spiegel, Fabian Mueller, Dorothea Wiesmann  \\
\{tep,fmu,dor\}@zurich.ibm.com \\ 
IBM Research Zurich \\
Saeumerstrasse 4, 8803 Rueschlikon, Switzerland \\ \\ 	
John Bird \\
johnbird@us.ibm.com \\
IBM TSS \\ 
Highway 52 North, 3605 Rochester, Minnesota, United States
}

\maketitle

\begin{abstract}
In predictive maintenance, model performance is usually assessed by means of precision, recall, and F1-score. 
However, employing the model with best performance, e.g. highest F1-score, does not necessarily result in minimum maintenance cost, but can instead lead to additional expenses.
Thus, we propose to perform model selection based on the economic costs associated with the particular maintenance application.
We show that cost-sensitive learning for predictive maintenance can result in significant cost reduction and fault tolerant policies, since it allows to incorporate various business constraints and requirements. 
\end{abstract}

\section{Introduction}

Predictive maintenance (PdM) uses machine learning models to forecast failures of mechanical or electronic devices that are prone to degradation \cite{Mobley02Predictive}.
Common use cases include wind turbines \cite{Canizo17Wind}, bearings \cite{Wu07Cost}, cylinder heads \cite{Gu17Cost} as well as medical equipment \cite{Sipos14Log}, servers subcomponents \cite{Botezatu16Disk, Giurgiu17Dram}, automated teller machines \cite{Wang17Atm}, and other IoT devices. 
Most PdM applications employ physical sensors \cite{Botezatu16Disk} or monitoring software \cite{Sipos14Log, Wang17Atm} or both \cite{Canizo17Wind, Giurgiu17Dram} to observe device behavior, which is subsequently consumed by the machine learning model that is responsible for making predictions. 

The objective is to learn degradation indicators or patterns that precede device failure \cite{Mobley02Predictive}, 
since recognizing these precursors allows for more efficient scheduling of maintenance and, consequently, increases the availability of the device \cite{Botezatu16Disk, Canizo17Wind, Giurgiu17Dram, Gu17Cost, Sipos14Log, Wang17Atm,  Wu07Cost}.
Training a PdM system actually corresponds to finding the model parameters that best fit or explain the observed device behavior. 
Best fit is usually defined by a cost function, which estimates the model performance by comparing actual and predicted condition for all devices.    
Since a device can be either broken or fully operational within a certain time interval, the PdM task is usually considered as a binary classification problem \cite{Freeman08Binary, Min09Binary}.
Cost functions for binary classification are commonly based on the confusion matrix and include the precision, the recall, and the F1-score \cite{Canizo17Wind, Giurgiu17Dram, Wang17Atm}.

However, the main shortcoming of these confusion-matrix-based performance measures is their lack of knowledge about the cost and effect of the PdM strategy.
Depending on the business case, important considerations for a practical PdM strategy are potential costs for ticket diagnosis and triage, field travel and repair, parts and shipping, as well as unavailability. 
Cost-sensitive learning with an economic cost function has been successfully applied to several condition-based PdM applications \cite{Deloux09Cost, Grall02Cost, Gu17Cost, Wu07Cost}.
Nonetheless, we see a gap between traditional cost functions for binary classification and economic cost functions for condition-based applications.   

Hence, we propose to incorporate all of the individual economic cost factors into the confusion matrix \cite{Elkan01Cost, Koyejo14Metric}, 
which produces an application specific cost function that allows us to optimize the PdM strategy, instead of maximizing decoupled performance measures.
Our experiments on a real-life PdM use case demonstrate that, in comparison to traditional optimization criteria, an application specific cost function is able to achieve significantly higher savings 
and yield fault tolerant policies.

The rest of the paper is structured as follows. 
Section \ref{sec:architecture} introduces a general PdM architecture, including the data processing, feature extraction, 
and model selection/evaluation step.    
The general PdM optimization problem and our application specific cost function are described in Section \ref{sec:optimization}.
Our experimental design and results are presented in Section \ref{sec:evaluation}, followed by a visual inspection of our cost function and a brief discussion of the potential savings in a real-life PdM use case.
A more general view on cost-sensitive learning and additional cost factors is provided in Section \ref{sec:generalized}.
We conclude with future work in Section \ref{sec:conclusion}.

\section{Predictive Maintenance}
\label{sec:architecture}

\begin{figure*}
\centering
\includegraphics[width=1.0\textwidth]{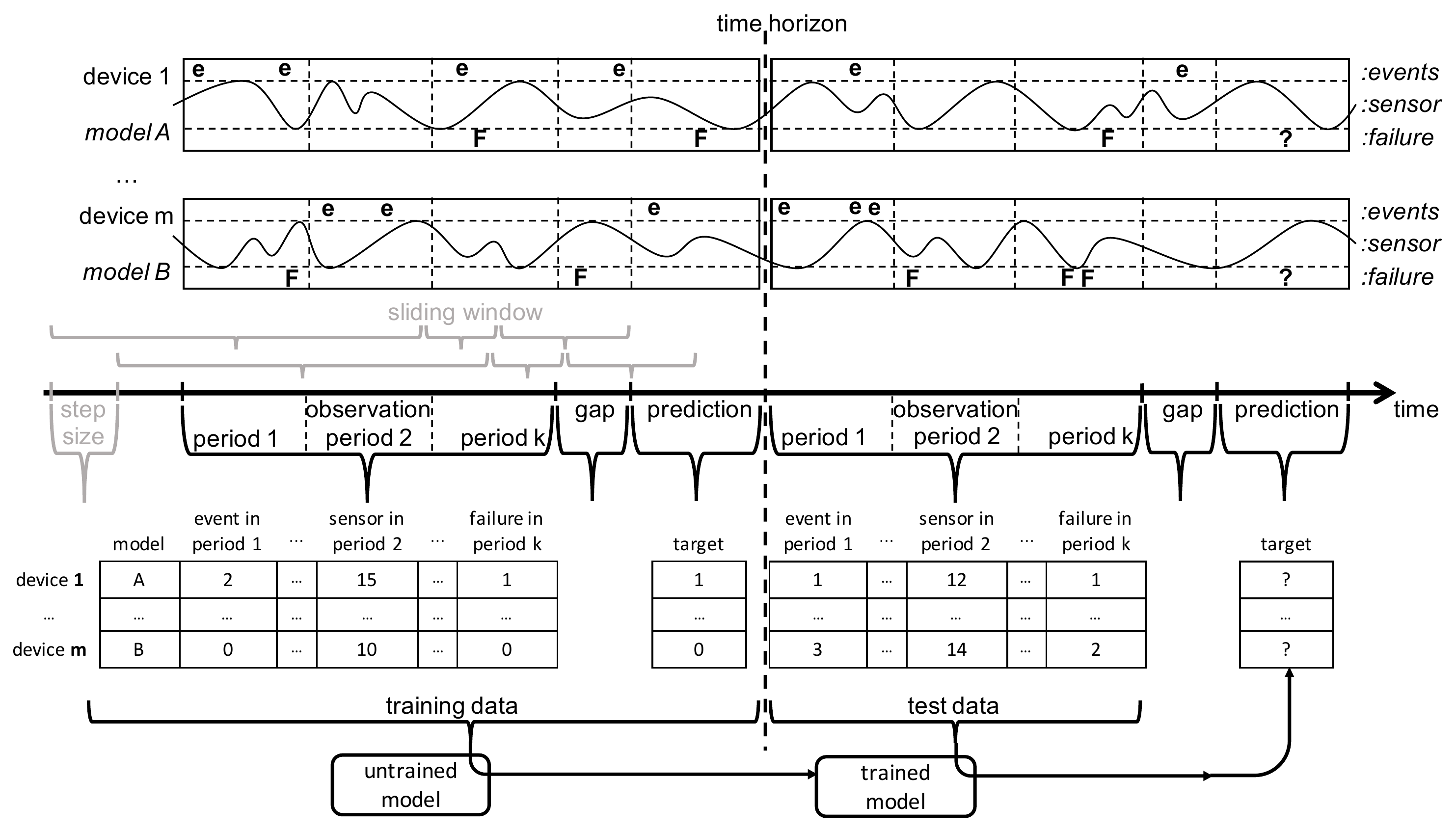}
\caption{Predictive maintenance architecture, illustrating our proposed data processing, feature extraction, and model evaluation for a set of sample device measurements.}
\label{fig:architecture}
\end{figure*}

Figure \ref{fig:architecture} shows the design of our proposed predictive maintenance architecture. In the following subsections we explain the illustrated data processing, feature extraction, and model selection/evaluation in more detail.

\subsection{Data Processing}

As presented in Figure \ref{fig:architecture}, we assume that we have observed behavior for a set of \textit{m} devices over a certain time interval. 
Behavioral data usually includes information about all kind of \textit{events} and \textit{failures} as well as \textit{sensor} measurements. 
For the purpose of model evaluation, we define a \textit{time horizon} (indicated by the dashed line that vertically intersects the time axis) that splits the temporal observations into a training and a test set respectively.

In order to create as many training examples as possible, 
the training set is often further divided by means of the sliding window technique, which examines different segments by moving along the time axis at a certain \textit{step size}. 
A \textit{sliding window} typically consists of an \textit{observation} and a \textit{prediction} interval, which are used to extract the model input and output (as exemplified by the feature matrix and target vector).
In some cases we furthermore define a \textit{gap} or rather a transition interval, which allows for sufficient time to schedule maintenance for the devices that are likely to fail within the prediction interval.
Please note that the \textit{step size} controls the amount of sliding window \textit{overlap}, see Figure \ref{fig:architecture}.

Similar PdM data processing architectures have been suggested for the failure prediction of memory modules \cite{Giurgiu17Dram} and automated teller machines  \cite{Wang17Atm}.

\subsection{Feature Extraction}
\label{sec:features}

Although there exist numerous techniques for modeling temporal behavior \cite{Canizo17Wind, Giurgiu17Dram, Grall02Cost, Wang17Atm, Wu07Cost}, we only describe the straight-forward (but very competitive) way of extracting temporal features and dependencies.
Given an \textit{observation interval}, we employ binning to extract a set of features for each of multiple consecutive time periods.
Figure \ref{fig:architecture} illustrates \textit{k periods}, each including an event/failure count and an average sensor value.

Although our example only includes low level features, it is very common to consider higher-features, such as mean time between failure or time elapsed since last failure \cite{Wang17Atm}, in order to generate more reliable predictions.  
 
Note that the output label or class of a device is determined by the absence or presence of a failure within the \textit{prediction interval}. 
Since the label can be either \textit{0} or \textit{1}, we consider predictive maintenance as a binary classification task.

\subsection{Model Selection}
Since we aim to capture temporal dependencies between individual bins or periods, it is recommended to employ a nonlinear model.
In literature we find several PdM systems that address the binary classification of device failures by means of gradient boosting and support vector machines \cite{Botezatu16Disk, Wang17Atm}, but also neural networks \cite{Wu07Cost} are not uncommon. 
A popular PdM classifier is the random forest model \cite{Botezatu16Disk, Canizo17Wind, Giurgiu17Dram, Wang17Atm}, which is also employed for our empirical evaluation (in Section \ref{sec:evaluation}).

The random forest model is advantageous, since it has only a few parameters and, therefore, requires comparatively little training data. 
Furthermore it can handle unbalanced data sets \cite{Chen04Rf}, allows for feature ranking, and is relatively easy to interpret.
Most importantly, the individual decision trees can model temporal dependencies, 
since their decision paths (root node to leaf node) assign an order to the features that were extracted from consecutive time periods 
(e.g.: more than one event in period 1 ---followed-by--$>$ average sensor reading of 15 in period 2 ---followed-by--$>$ at least on failure in period k ---results-in--$>$ device failure).

\subsection{Model Evaluation}

As shown in Figure \ref{fig:architecture}, our statistical model learns from one or more sliding windows that precede the defined time horizon.
The trained model is then used to predict failures for the test data by looking only at the observation interval and its corresponding features. 
Our hypothesis is that failure precursors are similar in training and test set, since both sets are drawn from the same distribution. 
Since we forecast the near future, we evaluate our model by setting the time horizon to the recent past (going back one sliding window length).
The discrepancy between predicted and actual failures gives us information about the model performance.

Common performance measures for binary classification problems (such as our predictive maintenance task describe above), include the precision, the recall, and the F1-score.
However, the following section will introduce other optimization criteria, which are often more practical in real-life PdM applications.

\section{Optimization Problem}
\label{sec:optimization}

This section discusses the traditional approach of model tuning and furthermore introduces a novel approach to derive a cost function that allows us to minimize maintenance expenses.

\subsection{Model Tuning}

We consider predictive maintenance as a binary classification problem, because for a given prediction interval a device can be categorized as either functional or inoperative. 

\begin{figure}[h]
\centering
\includegraphics[width=0.96\textwidth]{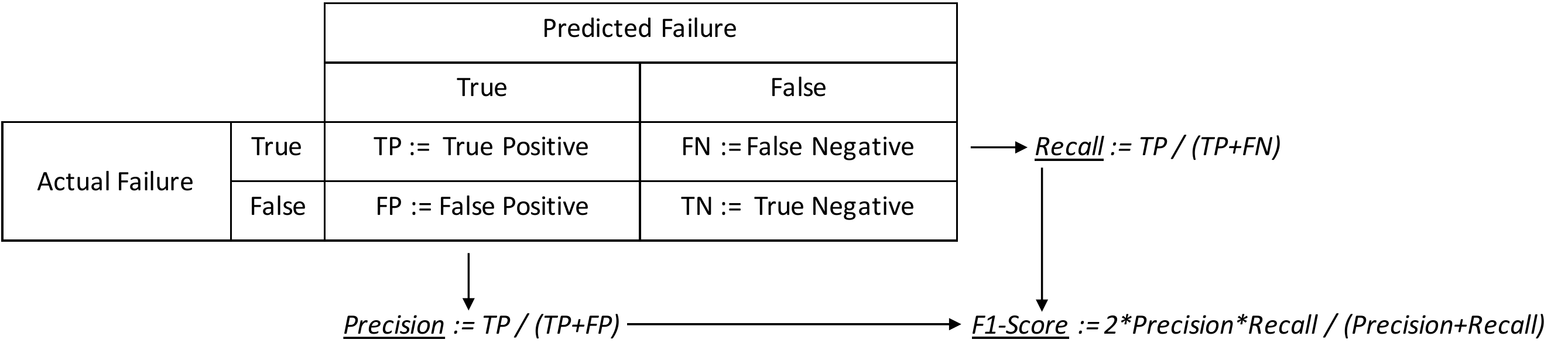}
\caption{Confusion matrix, illustrating the calculation of precision, recall, and F1-score.}
\label{fig:confusion}
\end{figure}

In general we aim at minimizing the error between predicted and actual failures by selecting the model parameters according a predefined performance measures. 
Commonly used measures include the precision, the recall, and their harmonic mean, the $F1$-score. 
Figure \ref{fig:confusion} illustrates a confusion matrix, which is used to calculate the $F1$-score and alternative performance measures.

Informally speaking, the recall can be interpreted as the percentage of actual failures that were correctly predicted, and the precision can be understood as the percentages of predicted failures that were actually true. 
Their harmonic mean, the $F1$-score, is widely used in the machine learning community to compare the performance of different models and parameter settings \cite{Koyejo14Metric}.

\begin{figure}[h]
\centering
\includegraphics[width=0.96\textwidth]{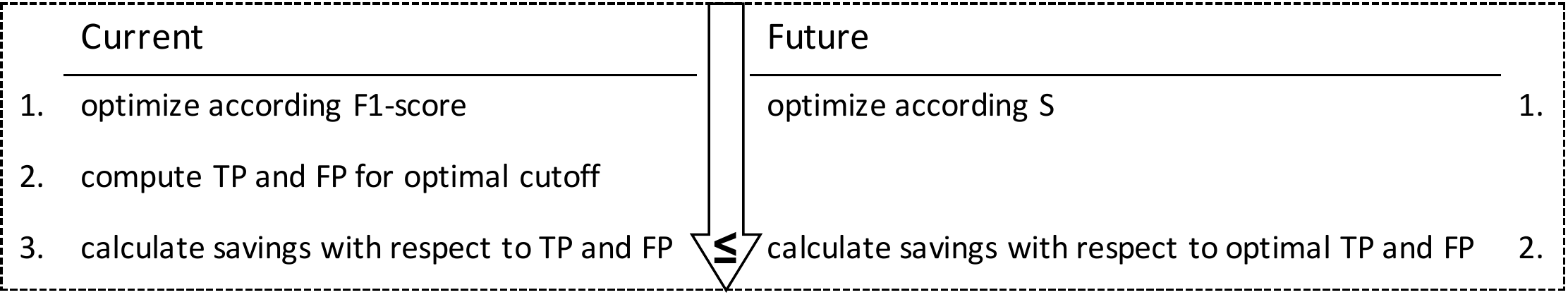}
\caption{Current and future approach of solving the PdM optimization problem. $S$ being our economic cost function.}
\label{fig:approach}
\end{figure}

Traditionally we aim at optimizing according to the $F1$-score, meaning that we select all model parameters in a way that the harmonic mean of precision and recall is maximized on the training set. 
One important parameter is the cutoff, which determines the threshold for binary classification and can be visualized as different points on the ROC curve. 
Given the optimal parameters, including the cutoff, we are in the position to calculate potential business savings of our trained model, refer to Figure \ref{fig:approach}.

However, for many predictive maintenance applications there is a mismatch between maximizing $F1$-score and minimizing maintenance cost. 
For that reason, we introduce an alternative approach of parameter selection, where the model is tuned in a way that it maximizes the savings $S$, as shown in Figure \ref{fig:approach}.
In the following we will explain how to derive an application-specific cost function, which can be used in the model parameter selection.

\subsection{Cost Function}

This subsection shows how to derive a custom cost function for an example business case, 
which uses \underline{realistic but artificial} costs that are inspired by a predictive maintenance case we have worked on.

\begin{figure}[h]
\centering
\includegraphics[width=0.86\textwidth]{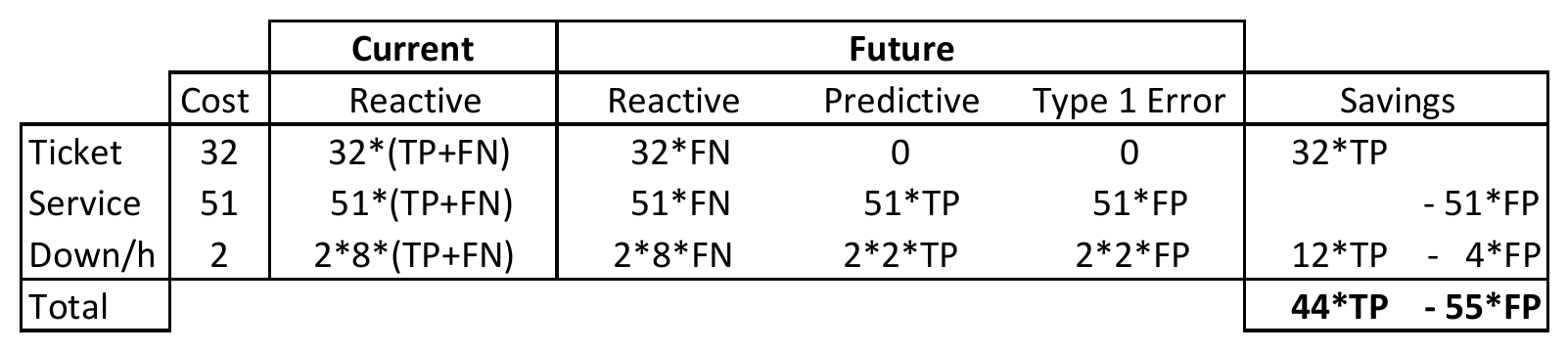}
\caption{Cost per incident. Savings are the delta between current and future approach. Downtime is billed per hour.}
\label{fig:cost}
\end{figure}

Considering a set of devices that are prone to degradation and consequently fail at times, we assume that each incident is associated with a certain cost.
Common cost factors include: (i) \textit{ticket} creation and processing, (ii) \textit{service} time and effort for repair or replacement of parts, and (iii) \textit{down} time during which the device is inoperable. 
Figure \ref{fig:cost} presents the approximate costs for our example use case, including the cost for reactive and predictive maintenance as well as the corresponding savings.

In case of reactive maintenance we account for $\$32$ ticket cost, $\$51$ service cost, and $\$16$ downtime cost ($6h$ travel and $2h$ repair) per incident.
Note that the total amount of incidents or actual failures is generally defined as the sum of true positives and false negatives, as illustrated by our confusion matrix above. 

In case of predictive maintenance we still account for reactive costs that were caused by false negatives, which represent the cases were a device failed, but our model predicted otherwise.
On the other side, for all correctly predicted outages or rather true positive we save ticket costs and downtime costs (for $6h$ field travel).
However, we need to account for additional costs that are created by false positives, where service and repair/replacement costs are produced for healthy devices that were falsely classified as problematic.

Figure \ref{fig:cost} presents the potential savings per incident, which are calculated by subtracting the future costs for predictive maintenance from the current costs for reactive maintenance.
In our example business case, the derived cost function for maximizing savings ($S = 44 {\cdot} TP {-} 55 {\cdot} FP$) gives an incentive to produce as many true positives and as few false positives as possible. 

As we will demonstrate by the experiments in Section \ref{sec:evaluation}, our newly derived cost function $S$ is able to achieve significantly higher savings, 
because (unlike the $F1$-score) it incorporates application-specific knowledge that benefits the optimization.

\section{Empirical Evaluation}
\label{sec:evaluation}

We are going to assess the performance of the discussed F1 and S cost function on a real-life predictive maintenance dataset. 
Our experimental design and results are described in the following subsections.
Moreover we present a visual inspection or comparison of both cost functions.

\subsection{Experimental Design}
\label{sec:evaluation:design}

In our experiments we consider behavioral data collected for a total number of $15'924$ devices. 
The dataset comprises temporal measurements over an interval of $6$ months, where week $1$-$12$ are utilized for training and week $13$-$24$ are employed for testing respectively. 
Training and test set both apply $10$ weeks observation, $1$ week transition, and $1$ week prediction interval. 
The resulting class imbalance for $1$ week prediction interval is approximately $1{:}2$, or more precisely $5633{:}10291$ and $5445{:}10479$ positive to negative cases for training and testing respectively. 

Given the described dataset, we have extracted $23$ categorical as well as $7$ numerical features. 
Categorical features include vendor, model, configuration, installation method, software version, geographical location, working hours, and other information about the devices. 
Numerical features contain failure statistics, such as the total number of incidents, the number of days that have passed since the last outage, and the mean time between failures. 
As explained in Section \ref{sec:features}, we employ binning to extract the discussed features for consecutive periods within the observation interval.

In our evaluation we employ the PdM dataset to demonstrate the influence of the cost function on the business savings that were achieved by the predictive model.
For experimentation we use the well-known random forest model, which is able to capture temporal dependencies between features that were extracted from consecutive observation periods.
The random forest model requires tuning of several hyper parameters, including the number of trees to grow (\textit{ntree}), 
the number of variables randomly sampled at each split (\textit{mtry}), the number of samples to draw (\textit{samp}), which is primarily used for datasets with class imbalance, 
and the \textit{cutoff} which decides if a test result is designated as positive or negative depending on whether the result value is above or below the threshold.
 
In general, the model parameters are tuned to produce optimal cost on a given training set. 
Since the parameter space is often large, we perform a grid search to evaluate only certain parameter settings, including:

\vspace{-0.33cm}
\begin{align}
ntree &= \{200, 400, 600, 800 \} \nonumber \\
mtry &= \{2, 3, 5, 7, 10, 15 \} = \lfloor \#features^ {\{0.3, 0.4, 0.5, 0.6, 0.7, 0.8 \}} \rfloor \nonumber \\
samp &= \{5633, 8447, \ldots, 10291\} = p * \{1.0, 1.5, \ldots \}, max(n) \nonumber \\
cutoff &= \{0.05, 0.06, \ldots, 0.95 \} \nonumber
\end{align}

As a rule of thumb, one usually sets \textit{mtry} to be the square root of the number of features. 
The \textit{samp} size can be a multiple of the number of positives $p$, as long as it does not exceed the number of negatives $n$.

\subsection{Experimental Results}

\begin{figure}[h]
\centering
\includegraphics[width=1.0\textwidth]{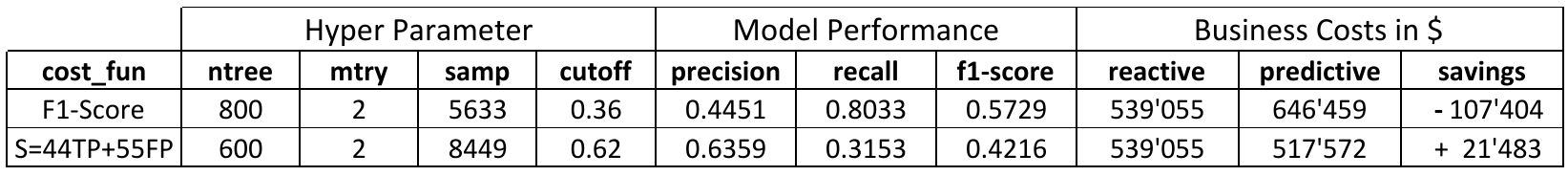}
\caption{Model parameter, performance, and savings for F1 and our cost function S.}
\label{fig:result}
\end{figure}

Given our PdM dataset, we have tuned the model parameters on the training set and evaluated the model performance on the test set. 
The selected model parameters, achieved model performance, and corresponding business savings, for both $F1$-score and our derived cost function $S$, are shown in Figure \ref{fig:result}.

Figure \ref{fig:result} illustrates that for both cost functions the \textit{ntree}, \textit{mtry}, and \textit{samp} parameters are set to similar values, but the \textit{cutoff} is selected quite differently. 
By taking a closer look at the model performance, we can see that the two cost functions place different emphasis on precision and recall.
Although our derived cost function S leads to less performance in terms of $F1$-score, it achieves much higher savings.
For almost $16K$ devices and $1$ week prediction interval, $S$ produced $\$21'483$ savings, 
whereas predictive maintenance with optimization according $F1$ created an additional cost of $\$107'404$ on top of reactive maintenance. 

\begin{figure}[t]
\centering
\includegraphics[width=0.68\textwidth]{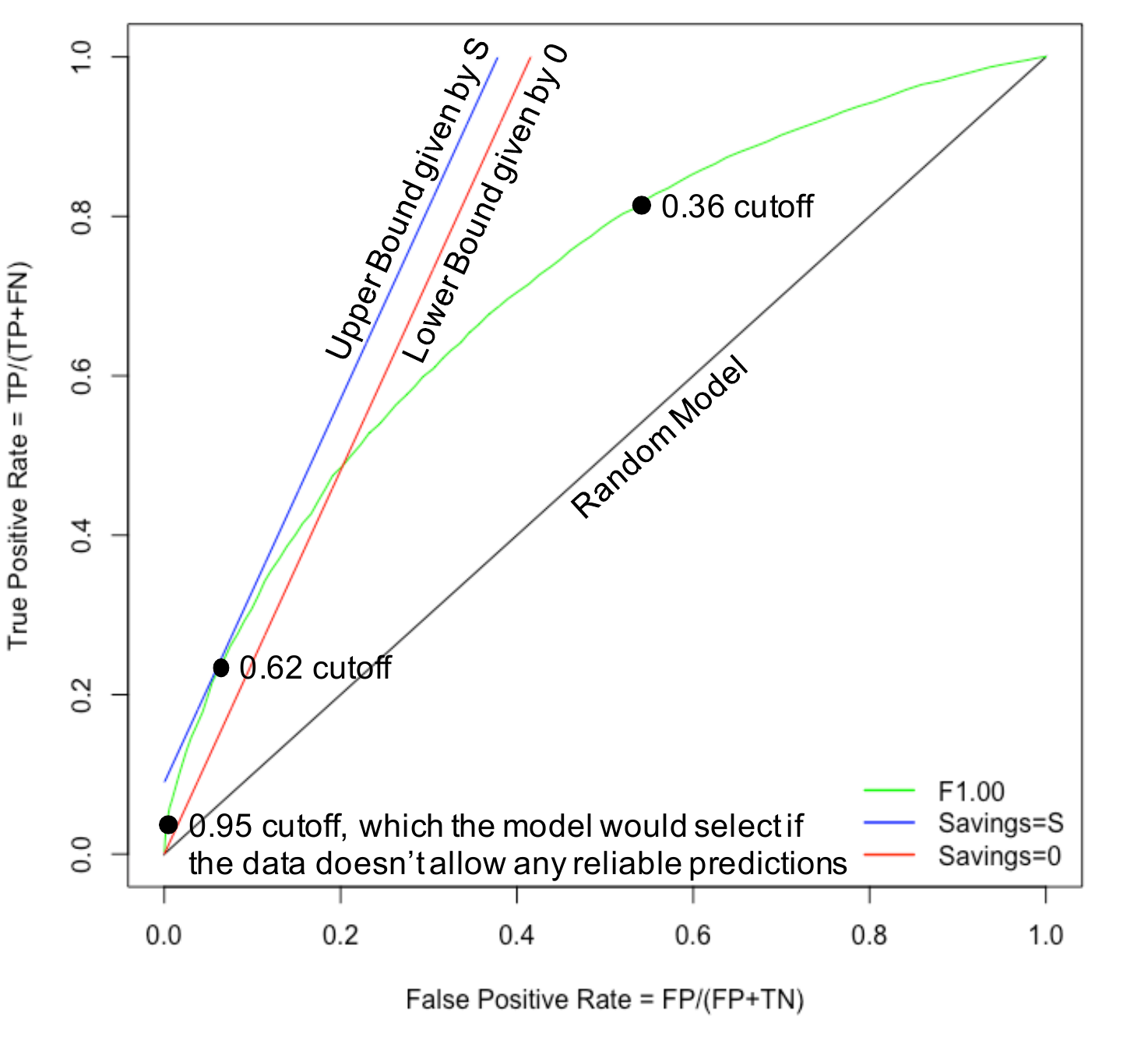}
\caption{ROC curve for F1 optimization, compared to both random model and upper/lower bound for savings S.}
\label{fig:roc}
\end{figure} 

Figure \ref{fig:roc} explains the difference in savings by illustrating the ROC curve for $F1$ optimization as well as showing the lower and upper bound of our derived cost function $S$.
In general, a ROC curve demonstrates the diagnostic ability of a binary classifier, plotting the true positive rate against the false positive rate, as the cutoff is varied. 
A random guess would give a point along the black diagonal line and the best possible prediction would yield in a point at the upper left corner or coordinate (0,1), representing no false positives and no false negatives. 
Furthermore, the bigger the area under the curve (AUC) the better the model, referring to its diagnostic ability to separate functional from inoperable devices.  

In Figure \ref{fig:roc}, the green ROC curve shows the performance of our random forest model for optimization according $F1$-score. 
The red and blue line illustrate the lower and upper bound for zero and $\$21'483$ savings, computed for all possible outcomes on our PdM dataset. 
We can see that the green ROC curve and the blue upper bound intersect at $0.62$ cutoff, which corresponds to the parameter setting that was selected by our derived cost function $S$. 
In contrast, optimization according $F1$-score yields $0.36$ cutoff, which is clearly above the random guess performance but way below the red zero savings line.
This illustrates that the cutoff parameter has a very strong influences on the savings  ($S = 44 {\cdot} TP {-} 55 {\cdot} FP$), since it controls the ratio between true and false positives. 

Figure \ref{fig:roc} highlights another important advantage of our derived cost function $S$, namely its \emph{fail-safe} property. 
Meaning that for low data quality our cost function would force the model to select a relatively high cutoff that corresponds to low risk in terms of false positives and low savings in terms of true positives. 
This \emph{passive mode} guarantees that $S \leq 0$, which is quite beneficial for online learning, where the model parameters are gradually updated according the newly incoming data.

\subsection{Visual Inspection}

\begin{figure}
\centering
\hspace{-1.3cm}
\includegraphics[width=0.88\textwidth]{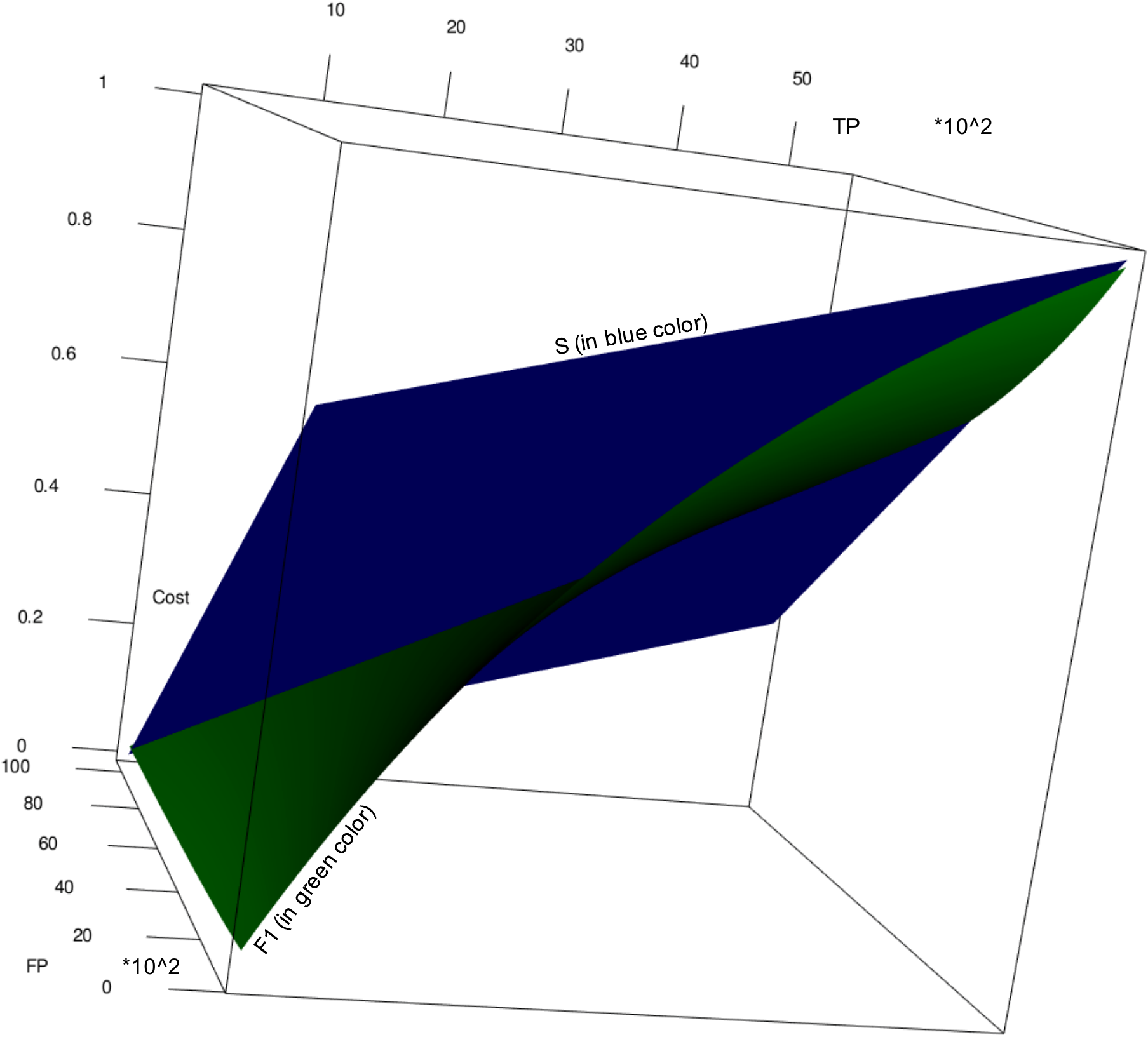}
\hspace{-1.4cm}
\text{(a)}
\vspace{0.6cm}
\mbox{}
\includegraphics[width=0.68\textwidth]{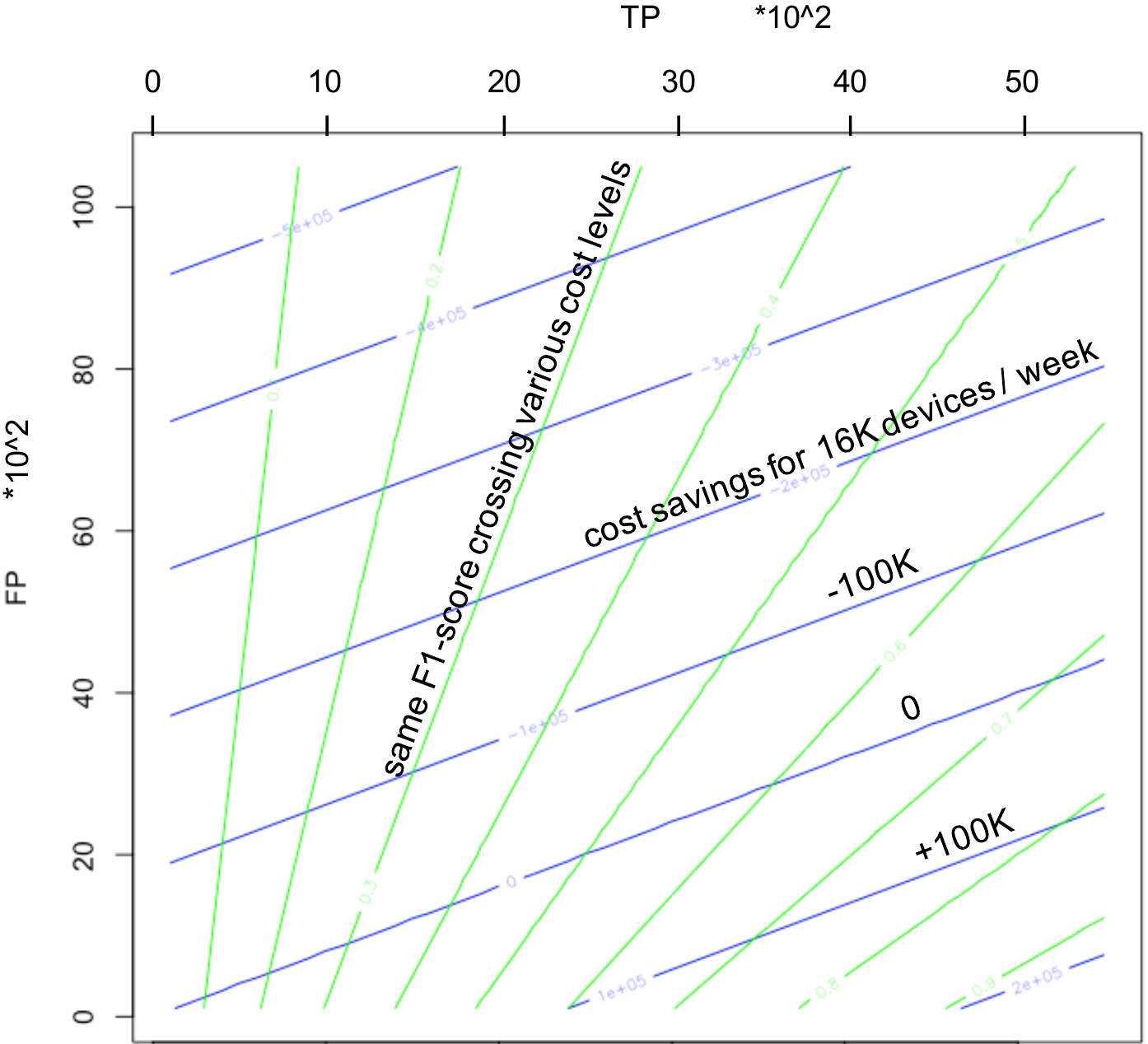}
\text{(b)}
\vspace{0.6cm}
\caption{(a) Hyperplane of F1 and S cost function, showing the normalized savings for all possible TP vs FP permutations. (b) Contour of F1 and S cost function, illustrating the absolute weekly savings (in \$) for all TP vs FP permutations.} 
\label{fig:visual}
\end{figure}

In order to achieve a better understanding of the examined cost functions, $F1$ and $S$, we compare their corresponding hyperplane and contour in Figure \ref{fig:visual}(a) and \ref{fig:visual}(b) respectively.

For the purpose of plotting the hyperplane and contour of both cost functions we need to compute $F1$ and $S$ for all values of $TP$ and $FP$. 
Since our test set contains $5445$ positive cases ($P {=} TP {+} FN$) and $10479$ negative cases ($N {=} FP {+} TN$), the number of permutations or data points ($P \times N \approx 57M$) is rather large. 
For our visual inspection we only consider $55{:}105$ cases, which gives us roughly the same positive:negative ratio as in the original test set but reduces the number of permutation by factor $10^4 = 10^2 \times 10^2$.  

In Figure \ref{fig:visual}(a) we compare the hyperplane of both cost functions by plotting the $F1$-score and normalized savings $S$ for all permutation of $TP$ and $FP$.
We see that the highest $F1$ and $S$ value is achieved for the maximum possible number of true positives and the minimum possible number of false positives.
Conversely, the minimum possible number of true positives and the maximum possible number of false positives resulted in the overall smallest $F1$ and $S$ value.  
However, we also see that the shape or curvature of the two hyperplanes is different, meaning that for a rather small number of true and false positives we get a relatively high $F1$-score but only small savings $S$. 

In Figure \ref{fig:visual}(b) we show the contour of both cost functions, which can be imagined as looking at the hyperplanes from a birds perspective. 
By taking a closer look at the blue iso-lines, we see that different permutations of true and false negatives lead to the same savings. 
For example, $\$100K$ cost savings for $16K$ per week can be achieved by around $24 \left( \cdot 10^2 \right) TP's$ and $0 \cdot \left( \cdot 10^2 \right) FP's$ 
as well as $55 \left( \cdot 10^2 \right) TP's$ and about $25 \cdot \left( \cdot 10^2 \right) FP's$.

Figure \ref{fig:visual}(b) furthermore illustrates that the green $F1$ iso-lines cross multiple cost levels, 
meaning that one and the same $F1$ score translates to $-100'000\$$, $0\$$, and $+100'000\$$ potential savings for $16K$ devices per week. 
This clearly shows that the $F1$-score is a ill-suited cost function, when it comes to business impact.

Although our actual classifier will probably only produce some of the possible $TP$ and $FP$ permutations, our visual inspection of the two cost functions shows their potential for real-world applications.

\subsection{Potential Savings}

In previous sections we have primarily discussed the monetary benefit of our derived cost function $S$, but there are still other factors that influence the potential savings.

First, the potential savings grow proportionally to the number of devices. 
For instance, if we apply our predictive maintenance solution to an application with twice as many devices, we would expect that savings have doubled.
In our previously discussed business case (refer to Figure \ref{fig:cost} and \ref{fig:result}), 
$16K {\times} 2$ devices result in $\$21'483 {\times} 2 {=} \$42'966$ savings, when optimizing according $S$.

Second, the potential savings grow linearly with the number of prediction windows.
Assuming that our predictive model is in production since a year and given that the failure rate is more or less constant, we would have saved $\$21'483\$ {\times} 53weeks {\approx} \$1.14M$,
which explains the interest in such solutions.

Furthermore, the potential savings strongly depend on the class imbalance, which in turn is influenced by the failure rate.
In general, relatively low failure rates diminish the potential savings, because underrepresented outages are harder to predict and, consequently, lead to less true positives and more false negatives.
For example, if a device fails only once in $26$ weeks, we would expect an approximate class imbalance of $1{:}25$ for $1$ week prediction interval.

Moreover, the class imbalance and, consequently, the potential savings are influenced by the length of the prediction interval. 
Even though smaller prediction windows cause higher class imbalance and pose a harder classification problem, 
there exist practical reason to avoid bigger prediction windows.
For instance, bigger prediction windows can result in less accurate failure times and too early repair or replacement of devices.

Finally, we want to mention that the potential savings furthermore depend on the gap or transition interval. 
For example, a $2$ week gap between the observation and prediction interval would make it much more difficult for our classifier to reliable predict outages, 
since we have to look further into the future, even though we lack information about how the devices will behave within the transition interval.
However, some applications require a transition interval in order to give the responsible service team enough time to react and schedule maintenance. 

Due to the fact that the transition and prediction interval play such an important role, we will present a more thorough analysis of these influencing factor in Section \ref{sec:generalized}.

\section{Economic Cost Function}
\label{sec:generalized}
Evaluating the prediction performance of PdM models based on custom cost functions allows us to properly address applications with varying requirements.
In the following we discuss some design principals that influence the potential savings.

\subsection{Evaluation of Model Performance}
Our predictive maintenance model computes a forecast $\hat{s}_j$ of the binary operational state $s_j$ (1 = in operation, 0 = out of operation) of a device $j$ during the prediction time interval $\st{T}_P$. Let ${\st{J}:=\{1,\dots,m\}}$, where $m$ is the total number of devices. Denote by
\begin{align}
	\tn{P} &:= |\{j\in\st{J}: s_j=1\}|,\\
	\tn{N} &:= |\{j\in\st{J}: s_j=0\}|
\end{align}
the number of devices that are in and out of operation during $\st{T}_P$, respectively. Note that ${\tn{N}+\tn{P}=\tn{J}}$. It is common to evaluate the predictive performance of a binary classification algorithm in terms of the number of true and false positives TP and FP, and true and false negatives TN and FN, respectively, defined as
\begin{align}
	\tn{TP} &:= |\{j\in\st{J}: \hat{s}_j=s_j=1\}|,\label{eq:defTP}\\
	\tn{FP} &:= |\{j\in\st{J}: \hat{s}_j=1 \land s_j=0\}|,\\
	\tn{TN} &:= |\{j\in\st{J}: \hat{s}_j=s_j=0\}|,\\
	\tn{FN} &:= |\{j\in\st{J}: \hat{s}_j=0 \land s_j=1\}|\label{eq:defFN}.
\end{align}
Because ${\tn{TN}=\tn{N}-\tn{FP}}$ and ${\tn{FN}=\tn{P}-\tn{TP}}$, it is sufficient to consider $\tn{TP}$ and $\tn{FP}$ only. Widely used performance measures include the recall RE, the precision PR, and the F1-score F1, which can be expressed as
\begin{align}
	\tn{RE}(\tn{TP}) &= \tn{TP}/\tn{P},\label{eq:defRE}\\
	\tn{PR}(\tn{TP},\tn{FP}) &= \tn{TP}/(\tn{TP}+\tn{FP}),\\
	\tn{F1}(\tn{TP},\tn{FP}) &= 2\tn{TP}/(\tn{TP}+\tn{FP}+\tn{P}).\label{eq:defF1}
\end{align}
The above performance measures can be informative with regard to classification quality. However, they might not be the most suitable measures for tuning the PdM model parameters in real-world applications where actual economic costs can be assigned to the quantities \eqref{eq:defTP}-\eqref{eq:defFN} explicitly.

\subsection{Generalized Cost Functions}

\begin{figure}
\centering
\includegraphics[width=0.48\textwidth]{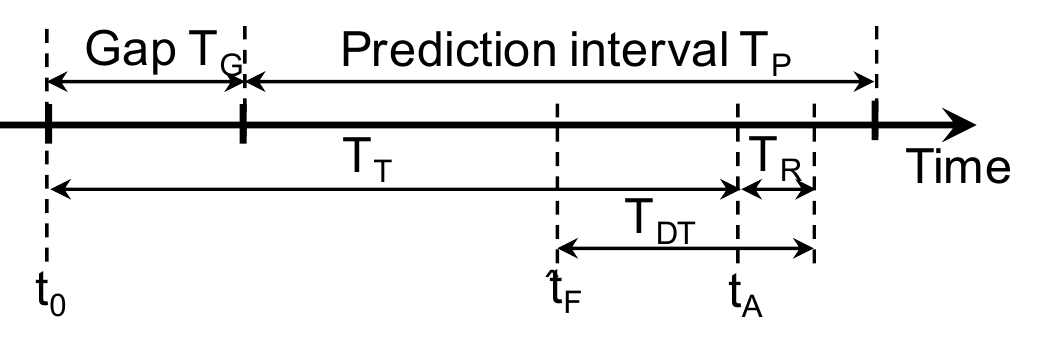}
\caption{Given the preparation and travel time $T_T$ and the repair time $T_R$, the total expected downtime $\hat{T}_{DT}$ depends on the gap duration $T_G$ and the prediction interval duration $T_P$.}
\label{fig:totaldowntime}
\end{figure}

It has been shown in Section \ref{sec:evaluation} that the performance of predictive maintenance models should be evaluated with regard to the actual cost structure of the application. In general, cost factors may include the incident ticket creation and processing, the service time and repair work, and the down time of the affected device. To properly model all the different cost structures of real-world applications, we propose to evaluate the performance of predictive maintenance models using a generalized performance measure $\pi$ selected from the family of real-valued functions ${\Pi: (\tn{TP},\tn{FP})\rightarrow\st{R}}$. Note that the performance measures \eqref{eq:defRE}-\eqref{eq:defF1} are comprised in $\Pi$. In particular, we are interested in affine performance measures of the form \\
\begin{equation}
	\pi(\tn{TP},\tn{FP}) = a(T_G,T_P)\tn{TP} + b(T_G,T_P)\tn{FP} + c(T_G,T_P),
	\label{eq:defPi}
\end{equation}
where ${a(T_G,T_P),b(T_G,T_P): \st{R}^2\rightarrow\st{R}}$ denote the costs per true positive TP and false positives FP, respectively, and can be arbitrary functions of the gap $T_G$ and the prediction interval $T_P$, as illustrated in Figure \ref{fig:totaldowntime}. The offset ${c(T_G,T_P): \st{R}^2\rightarrow\st{R}}$ can also depend on $T_G$ and $T_P$.

To illustrate why the cost coefficients $a$ and $b$ can depend on $T_G$ and $T_P$, we revisit the predictive maintenance use-case described in Sections \ref{sec:optimization} and \ref{sec:evaluation}. 
The predictive maintenance model is executed at time $t_0$ to predict if for a particular device there will be a failure during the prediction interval ${\mathbb{T}_P:=[t_0+T_G,t_0+T_G+T_P]}$, see Figure \ref{fig:totaldowntime}. However, the model does not provide any information on the exact time $t_F$ the failure will occur. Thus, we consider the time of a failure as a random variable uniformly distributed on the prediction interval, \ie, ${t_F\sim\mathcal{U}(\mathbb{T}_P)}$. The expected time of failure is ${\hat{t}_F:=\mathbf{E}[t_F]=t_0+T_G+T_P/2}$. 
If, at time $t_0$ the model predicts a failure for a particular device, the operator orders the required spare parts, plans the repair work, and sends off the repair personnel. The earliest possible time for the personnel to arrive at the device location is $t_A:=t_0+T_T$, where $T_T$ summarizes the time required for preparations and travel. A device failure occasions different types of costs as summarized in Table \ref{fig:cost}. One cost component is the cost of the device downtime. The expected downtime $\hat{T}_D$ of a device depends on the gap duration $T_G$ and prediction interval length $T_P$ and is computed as
\begin{align}
	\hat{T}_D(T_G,T_P)&=\max(0, t_A-\hat{t}_F)+T_R\nonumber\\
	&=\max(0,T_T-T_G-T_P/2)+T_R,
	\label{eq:expectedDT}
\end{align}
where $T_R$ denotes the repair time. That is, if the repair personnel arrives before the expected time of failure, the total downtime equals the repair time. If, however, the personnel arrives only after the failure, the time difference $t_A-\hat{t}_F$ must be added to the repair time. The downtime costs per device failure are $2\hat{T}_D(T_G)$ \$.

While repairing a device after a failure results in additional downtime costs, executing the repair work and replacing device components long before they fail is also not desirable because it needlessly reduces their operating life. We assume that the value of a device component is equally distributed over its expected life time, \ie, the value of a device per operating time equals $C/\hat{T}_L$, where $C$ and $\hat{T}_L$ are the procurement cost and expected life time of the component, respectively. Consequently, if a device component is replaced $\Delta T$ time units before its failure, the fraction $\Delta T/\hat{T}_L C$ of the component's value is lost. Thus, the optimal repair strategy is to schedule the repair of a device at its expected failure time $\hat{t}_F$ so that no value is lost both in the reactive case and in the case of a true positive. However, in case of a false positive, a device component is replaced at time $\hat{t}_F$ even though it would not have failed during this prediction horizon resulting in a loss of at least $T_PC/(2\hat{T}_L)$ for every falsely predicted failure. For our example we have ${C/\hat{T}_L\approx0.12}$ \$/h.

\begin{figure}
\centering
\includegraphics[width=0.8\textwidth]{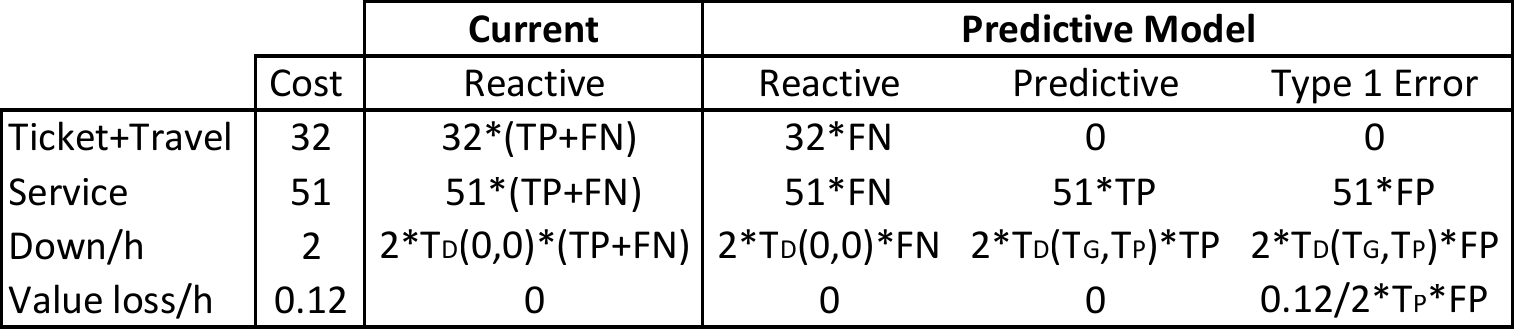}
\caption{Generalized cost components depending on the gap duration $T_G$ and the prediction interval length $T_P$.}
\label{fig:genCosts}
\end{figure}

Figure \ref{fig:genCosts} summarizes the generalized cost components. In contrast to the original formulation in Figure \ref{fig:cost}, the downtime costs take into account the expected downtime according to \eqref{eq:expectedDT}, where the reactive case corresponds to ${T_G=0}$ and ${T_P=0}$. An additional cost component has been introduced to penalize the premature replacement of device components. Based on this cost structure, the monetary savings $S$ resulting from the use of a predictive model are
\begin{align}
	S(TP,FP) =&\, [32+2(\hat{T}_D(0,0)-\hat{T}_D(T_G,T_P))]\tn{TP}\nonumber\\
	& - [51+2\hat{T}_D(T_G,T_P) +0.06T_P]\tn{FP},
	\label{eq:genSavings}
\end{align}
which has the form \eqref{eq:defPi}.

Choosing a shorter gap duration and a longer prediction interval in general improves the prediction performance of our PdM model, and, simultaneously, increases the downtime costs and component value loss discussed above. Varying the parameters $T_P$ and $T_G$ influences the structure of the savings \eqref{eq:genSavings} as well as the model's prediction accuracy, \ie TP and FP. Table \ref{tab:genSavings} summarizes the savings obtained for different gap and prediction interval durations for the cases where \textit{i)} the F1-score \eqref{eq:defF1} and \textit{ii)} the actual savings \eqref{eq:genSavings} were used as the performance measure based on which the best PdM model is determined. Consider for example the base case ${(T_G=7,T_P=7)}$. The reactive maintenance costs for all devices over the prediction interval ${T_P=7}$ d is 539055\$. If a PdM model selected based on the F1-score is used, additional costs of 109694\$ (20.35\%) are produced mainly due to false positives. If, however, the PdM model is selected based on the custom economic cost function $S(TP,FP)$, \cf\eqref{eq:genSavings}, the maintenance costs can be reduced by 21070\$ (3.91\%). These results suggest that it is economically advantageous to use the actual cost function $S(TP,FP)$ rather than a generic quality measure such as the F1-score for the model selection.

\begin{table}
\centering
\resizebox{0.86\textwidth}{!}{
\begin{tabular}{|c|c|c|c|c|}\hline
$(T_G,T_P)$ & Reactive & PdM F1 & PdM S & $\Delta$ \\ \hline
 (4,4) & \$346'302=100\% & \$462'682=133.61\% & \$343'559=99.21\%  & 34.40\% \\
 (4,7) & \$534'699=100\% & \$610'392=114.16\% & \$516'475=96.59\%  & 17.56\% \\
 (4,10) & \$653'598=100\% & \$692'966=106.02\%  & \$610'696=93.44\% & 12.59\% \\ \hline
 (7,4) & \$388'476=100\% & \$525'105=135.17\% & \$384'685=99.02\%  & 36.15\% \\
 (7,7) & \$539'055=100\% & \$648'749=120.35\%  & \$517'985=96.09\% & 24.26\% \\
 (7,10) & \$674'487=100\% & \$714'741=105.97\% & \$623'421=92.43\% & 13.54\% \\ \hline
 (10,4) & \$350'163=100\% & \$462'652=132.12\% & \$345'976=98.80\% & 33.32\% \\ 
 (10,7) & \$532'521=100\% & \$617'586=115.97\% & \$511'966=96.14\% & 19.83\% \\
 (10,10) & \$666'963=100\% & \$709'105=106.32\% & \$617'779=92.63\% & 13.69\% \\ \hline
\end{tabular}
}
\caption{Evaluation of the economic savings of using our PdM model for different gap and prediction intervals $T_G$ and $T_P$, respectively, see (\ref{eq:genSavings}).}
\label{tab:genSavings}
\end{table}

\section{Conclusion and Future Work}
\label{sec:conclusion}

There exists a great potential for predictive maintenance (PdM), since the number of sensor-equipped devices and the need for effective maintenance strategies is growing. 
However, we have discovered a mismatch between the PdM performance criteria and the business requirements.
Traditional optimization criteria, such as the F1-score, favor PdM models that correctly forecast a high number of failures, but they usually neglect the economic cost associated with true/false positive/negatives.   

We propose to closely examine the business processes in order to gain a better understanding of the cost structure and incorporate the individual cost factors into the PdM optimization.
An application-specific cost function has been introduced as well as compared to traditional performance measures.
Our evaluation has demonstrated that the proposed cost function is able to achieve significantly higher savings and furthermore prevents financial loss caused by inaccurate predictions on low quality data.

In addition we have presented a general recipe for integrating various business objectives into an economic cost function, which is achieved by assigning weights to the individual components of a confusion matrix.
Moreover, our study demonstrates that it is possible to design cost functions that incorporate general PdM objectives, such as a long transition and short prediction window.
We believe that our proposed recipe has the potential to provide even better PdM solutions in many application area.

In the near future we will deploy our solution in a production environment and extent it to different PdM use cases. 
Furthermore, we will investigate additional cost factors, such as the cost of different service personal involved in the maintenance process.
Another interesting and untouched aspect is the influence of the sliding window step size and the potential of overlapping prediction windows.

\end{document}